\documentclass{article}
\usepackage{ijcai20}

\usepackage{times}
\usepackage{soul}
\usepackage{url}
\usepackage[hidelinks]{hyperref}
\usepackage[utf8]{inputenc}
\usepackage[small]{caption}
\usepackage{graphicx}
\usepackage{amsmath}
\usepackage{amsthm}
\usepackage{booktabs}
\usepackage{algorithm}
\usepackage{algorithmic}
\usepackage{makecell}
\setcellgapes{3.3pt}
\usepackage{footnote}
\makesavenoteenv{tabular}
\makesavenoteenv{table*}
\usepackage{kotex}
\usepackage{float}
\usepackage{textcomp}
\usepackage[bottom]{footmisc}

\title{KR-BERT\footnote{\url{https://github.com/snunlp/KR-BERT}}: A Small-Scale Korean-Specific Language Model}

\author{Sangah Lee, Hansol Jang, Yunmee Baik, Suzi Park, Hyopil Shin \\ \affiliations{Seoul National University}}

\begin{document}

\maketitle

\begin{abstract}
    Since the appearance of BERT, recent works including XLNet and RoBERTa utilize sentence embedding models pre-trained by large corpora and a large number of parameters. Because such models have large hardware and a huge amount of data, they take a long time to pre-train. Therefore it is important to attempt to make smaller models that perform comparatively. In this paper, we trained a Korean-specific model KR-BERT, utilizing a smaller vocabulary and dataset. Since Korean is one of the morphologically rich languages with poor resources using non-Latin alphabets, it is also important to capture language-specific linguistic phenomena that the Multilingual BERT model missed. We tested several tokenizers including our BidirectionalWordPiece Tokenizer and adjusted the minimal span of tokens for tokenization ranging from sub-character level to character-level to construct a better vocabulary for our model. With those adjustments, our KR-BERT model performed comparably and even better than other existing pre-trained models using a corpus about 1/10 of the size.
\end{abstract}

\section{Introduction}

The Bidirectional Encoder Representations from Transformers (BERT) Multilingual model \cite{devlin2018bert} were pre-trained on Wikipedia data for 104 languages and has demonstrated powerful performances on a wide range of tasks in Natural Language Processing. Although it is meant to act as a universal language model, it does not always work well for non-English downstream tasks. This is due to the fact that it cannot fully capture the specific linguistic characteristics of every language. To accommodate for this, we have prepared a single language model that incorporates this knowledge for Korean.

Developing a BERT model for Korean NLP tasks is challenging for two reasons. First, Korean is an agglutinative language, which is morphologically richer than inflectional languages such as German or French. Secondly, its writing system, Hangul, is composed of more than 10,000 syllable characters. For these reasons, a proper BERT model for Korean requires both a vocabulary and a tokenizer that can effectively treat a variety of complex word forms.

Additionally, the scale of the multilingual BERT model is too large to be applied to single-language tasks. The model is trained on data combining texts from 104 languages and utilizes a vocabulary including characters from those 104 languages, which results in larger number of parameters. While there are other existing large-scale single language models such as XLNet or RoBERTa, such large models require a huge amount of training data and resources, which makes it difficult to train them for languages with fewer resources. Therefore it is important to scale the model down while still maintaining comparable performances, as recent works such as ALBERT and DistilBERT do.

In this paper, we construct our small scale single language model by utilizing a smaller number of parameters and vocabulary entries, as well as using less training data. We built a Korean BPE vocabulary with sub-characters as the smallest units instead of characters, and propose the BidirectionalWordPiece Tokenizer to capture the characteristics of the Korean language. Our KR-BERT model, which we trained from scratch on our own corpus, outperforms the BERT-base Multilingual model on four downstream tasks, specifically sentiment analysis, Question-Answering, Named Entity Recognition (NER), and Paraphrase Detection. The model shows the most consistent results and is comparable to other Korean-language BERT models such as KorBERT and KoBERT, despite its small scale.

\section{Related Work}
\subsection{Models after BERT}
Since the powerful performance that BERT demonstrated on a wide range of tasks in NLP, various improved BERT models have been shown. There is a trend towards bigger models such as XLNet and RoBERTa which utilize a larger corpora than the 16GB BERT and a large number of parameters. On the other hand, smaller models such as DistilBERT and ALBERT are implemented to overcome hardware limitations and longer training times of the huge models. \autoref{tab:overview} shows a quick overview of existing models mentioned above.

\begin{table*}[t]
    \makegapedcells
    \centering
    \resizebox{\textwidth}{!}{
    \begin{tabular}{l|ccccc}  
    \toprule
    & \textbf{BERT} & \textbf{XLNet} & \textbf{RoBERTa} & \textbf{DistilBERT} & \textbf{ALBERT} \\
    \midrule
    \textbf{\makecell[l]{The \# of \\ Parameters (M)} } & \makecell{Base: 110 \\ Large: 340} & \makecell{Base: ~110 \\ Large: ~340} & \makecell{Base: 110 \\ Large: 355} & Base: 66  & \makecell{Base: 12 \\ Large: 18} \\  \midrule
    \textbf{Data} & \makecell{16GB BERT data \\ 3.3B words} & \makecell{13GB BERT data \\ + over 130GB \\ additional data \\ 33B words} & \makecell{BERT data \\ + over 140GB \\ additional data} & BERT data & BERT data \\  \midrule
    \textbf{Method} & \makecell{BERT \\ (Bidirectional \\ Transformer with \\ MLM \& NSP)} & \makecell{Bidirectional \\ Transformer \\ with Permutation \\ based modeling} & BERT without NSP & BERT Distillation & \makecell{Factorized Embedding, \\ Cross-layer \\ Parameter Sharing, \\ SOP} \\
    \bottomrule
    \end{tabular} 
    }

    \caption{An overview of existing BERT-based models. MLM, NSP, and SOP in the table stand for the tasks related to BERT: Masked Language Modeling, Next Sentence Prediction, and Sentence Order Prediction.}
    \label{tab:overview}
\end{table*}

\begin{table*}[t]
    \makegapedcells
    \centering
    \resizebox{\textwidth}{!}{
    \begin{tabular}{l|cccccc}  
    \toprule
    & \textbf{KorBERT} & \textbf{KoBERT} & \textbf{\makecell{Kakao NLP Team \\ (Park, 2018)}} & \textbf{\makecell{KoreanCharacter\\BERT}} & \textbf{\makecell{KalBert \\ based on ALBERT}} & \textbf{\makecell{DistilKoBert \\ (based on DistilBERT)}} \\
    \midrule
    \textbf{Tokenizer} & \makecell{morphome-level \\ and character-level \\ (WordPiece)} & \makecell{character-level \\ (SentencePiece)} & morpheme-level & \makecell{character-level \\ (modified WordPiece)} & \makecell{morpheme-level \\ (BPE) wihtout tag} & \makecell{character-level \\ (SentencePiece)} \\
    \textbf{Data} & 23GB & \makecell{25M sents, \\ 324M words} & \makecell{Munjong Corpus} & \makecell{its own \\ Korean Dataset} & \makecell{6GB} & \makecell{6GB} \\
    \textbf{\makecell[l]{Additional \\ Information}} & \makecell{Details below} & \makecell{Details below} & - & \makecell{7477 vocab \\ Small: 3 layers \\ Base: full layers} & 47471 vocab & \makecell{3 layers \\ (reducing 12 layers \\ of KoBERT)} \\
    \bottomrule
    \end{tabular} 
    }
    \caption{An overview of existing Korean-specific models}
    \label{tab:kormodels}
\end{table*}

\subsection{Language-specific BERT models}

Multilingual BERT is pre-trained on Wikipedia data for 104 languages. Although the multilingual BERT shows remarkable cross-lingual ability, a variety of single-language BERT models are suggested for improvement. Language-specific BERT models are trained on German\footnote{\url{https://deepset.ai/german-bert}}, Italian \cite{PolignanoEtAlCLIC2019}\footnote{\url{https://github.com/marcopoli/AlBERTo-it}}, Finnish \cite{virtanen2019multilingual}\footnote{\url{https://github.com/TurkuNLP/FinBERT}}, and Japanese \cite{bertjapanese}\footnote{\url{https://github.com/yoheikikuta/bert-japanese}}. In addition, the CamemBERT model for French \cite{martin2019camembert} is a RoBERTa model \cite{liu2019roberta} that was trained on a French corpus of a similar size to the training data of the multilingual BERT, and the Chinese BERT model trained on a Chinese Wikipedia is a combination of BERT, BERT with Whole Word Masking, and RoBERTa models \cite{cui2019pre}\footnote{\url{https://github.com/ymcui/Chinese-BERT-wwm}}.

\subsection{Korean}
\subsubsection{Subword Segmentation and Tokenization}
\paragraph{BPE and its Applications.} A typical BERT model uses either the WordPiece tokenizer \cite{wu2016google} or the SentencePiece tokenizer \cite{kudo-2018-subword}, which are both based on the Byte-Pair-Encoding (BPE) algorithm \cite{sennrich-etal-2016-neural}.

\paragraph{Language-specific Vocabulary and Linguistic Structure.} Kim \textit{et al.} \shortcite{kim2019advanced} suggest the use of a subword segmentation algorithm in order to reflect the linguistic structure of Korean. They utilize Korean linguistic characteristics such as particles and normalize subwords based on mutual information.

Possibilities for tokenization of Korean include morpheme-level, character-level, and sub-character level models, however, no sub-character level model has been presented so far.

\subsubsection{Recent Korean BERT models}
Recent Korean BERT pre-training models that have been openly released include KorBERT (Electronics and Telecommunications Research Institute 2019)\footnote{\url{http://aiopen.etri.re.kr/service_dataset.php}} and KoBERT (SK Telecom 2019)\footnote{\url{https://github.com/SKTBrain/KoBERT}}. The details of these models are examined in sections 4 and 5. Park \shortcite{park2018bert}\footnote{\url{http://docs.likejazz.com/bert}} reports that the Kakao NLP Team is pre-training a BERT model with a Korean corpus. Some Korean models are presented on Github: KoreanCharacterBert, KalBert and DistilKoBert. \autoref{tab:kormodels} shows the overall comparison of those models.

\section{The Need for a Small-scale Language-specific Model}

The multilingual BERT model shows weak performance dealing with the Korean texts considering its scale. Also, the language-specific BERT models trained on other languages show better performances than multilingual BERT model on downstream tasks. We analyze the disadvantages of the multilingual BERT model below.

\subsection{Limit of Corpus Domain}
While the multilingual BERT model was trained on Wikipedia texts from 104 languages, the language-specific BERT models like German BERT or French CamemBERT were trained on several diverse data sources, including legal data, news articles, and so forth. The multilingual BERT is therefore limited in its domain with respect to language usage, since Wikipedia articles have different linguistic properties from other user-generated content, such as blogs or user comments \cite{ferschke2014quality,habernal2017argumentation}. This weakness is especially notable for noisy user-generated texts. To solve this problem, we can expand the variety of data sources that the model is trained on.

\subsection{Considering Language-specific Properties}
\subsubsection{Rare ``Character'' Problem}
In the process of subword tokenization of the transformer mechanism, misspelled or rare words can be handled for languages written in Latin alphabets such as English by simply separating the word into letters. For example, a token such as `lol' which is not present in the multilingual BERT vocabulary can be tokenized into subword units such as l \#\#o \#\#l. This is possible because the 26 characters of the Latin alphabet can be included in the whole vocabulary without significantly increasing vocabulary size of the model. In contrast, a Korean word such as 힉 hik ``eek'' can not be decomposed in the same manner due to the whole word being treated as a single character, not present in the multilingual BERT vocabulary. Because Korean is organized by syllable and not by letter, augmenting the vocabulary, as is done with the Latin alphabet, would greatly increase the vocabulary size. 

There are a total of 11,172 existing syllables in the written system of the language. However, only 1,187 syllables out of them are included in the vocabulary of multilingual BERT, which means that the excluded 9,985 syllables can not be analyzed properly. For this reason, we need a new BERT vocabulary that is able to solve this problem in order to improve results on Korean NLP tasks.

\subsubsection{Inadequacy for Morphologically Rich Languages}
Because Korean is an agglutinative language with generally one feature encoded per suffix, the morphology is much richer than English. Fusional languages with grammatical affixes potentially encoding several different features, such as French or German, also have more complex morphology than English. However, for an agglutinative language there will be greater difficulties in representing the vocabulary because of its morphological complexity. Specifically, there is a lack of representation for verb conjugations. Unlike for English, in languages such as Japanese and Korean, each verb conjugation can have dozens of different forms.

\subsubsection{Lack of Meaningful Tokens}
For the German language, it was found that the vocabulary of the multilingual BERT model contains a large number of individual subword units that do not have a clear semantic meaning. A similar type of problem is present for Korean, in which most words are tokenized as single characters, rather than morpheme-like units. This signifies that the multilingual model would have difficulty with properly representing multiple languages.

To solve these problems, we implemented the language-specific vocabulary and tokenizer which can deal with Korean texts and the language-specific properties of Korean. We propose two levels of vocabularies: syllable character and sub-character. One Korean character, which corresponds to one syllable, can be decomposed into smaller sub-character units, or graphemes. It is possible to construct the vocabulary more efficiently using sub-characters than cramming all the 11,172 syllables into the vocabulary.

\subsection{Large Scale of the Model}
Existing bigger models utilize very large amounts of data for training: the multilingual BERT model collected Wikipedia texts from 104 languages, XLNet used 130GB and RoBERTa used 160GB of data. They also have a large number of parameters with multilingual BERT utilizing 167M and RoBERTa using 355M parameters. These large-scale models require a huge amount of resources in order to perform. We hope to maintain comparable performances using a smaller vocabulary, fewer parameters, and less training data.

\section{Models}
In order to resolve the problems discussed above, we design the Korean-specific KR-BERT model, with the process described below. It has a relatively small data size for pretraining, but still performs comparably to the existing Korean-specific models. We suggest both character and sub-character models with two tokenizers: BERT original tokenizer and our BidirectionalWordPiece tokenizer. Our new vocabulary and BidirectinalWordPiece tokenizer are designed to reflect features of Korean language.

\subsection{Subcharacter Text Representation}
Korean text is basically represented with Hangul syllable characters, which can be decomposed into sub-characters, or graphemes. To accommodate such characteristics, we trained a new vocabulary and BERT model on two different representations of a corpus: syllable characters and sub-characters. When the BPE algorithm is applied to each of these two representations, the resulting tokenization can be different. For example, a syllable character 뜀 ttwim ``jumping'' can be decomposed into two sub-character units ㄸㅟ ttwi ``jump'' and ㅁ m ``-ing'' \cite{park2018grapheme}.

Unlike KorBERT and KoBERT, the KR-BERT model uses a sub-character representation in addition to a character representation. This method has the advantage of allowing us to capture the common aspects of various Korean verb/adjective conjugation forms, since many grammatical morphemes are realized only by partial elements of a syllable character. 

As seen in \autoref{tab:korrepr}, a sub-character representation can detect these elements without a morphological analyzer, while the character representation cannot do so. It can capture what the multilingual BERT cannot detect.

\begin{table}[t]
    \makegapedcells
    \centering
    \resizebox{\columnwidth}{!}{
    \begin{tabular}{ll}  
    \toprule
    \textbf{Character representation (default)} & \textbf{Sub-character representation (decomposed)} \\
    \midrule
    갔 kass ``went'' & ㄱk ㅏa ``go'' ㅆss ``-ed'' \\
    감 kam ``going'' & ㄱk ㅏa ``go'' ㅁm ``-ing'' \\
    간 kan ``that has/have gone'' & ㄱk ㅏa ``go'' ㄴn ``that has/have \ldots'' \\
    갈 kal ``that will go'' & ㄱk ㅏa ``go'' ㄹl ``that will \ldots'' \\
    \bottomrule
    \end{tabular} 
    }
    \caption{Two representation methods of the verb conjugations of 가 ka ``go''}
    \label{tab:korrepr}
\end{table}

\subsection{Subword Vocabulary}
We used the BPE algorithm to obtain a fixed-size vocabulary. Different vocabulary sizes were experimented with, ranging from 8,000 to 20,000. For each vocabulary size, we measured Masked LM Accuracy at 100,000 steps for the BERT pre-training stage. The best results were obtained with a vocabulary size of 10,000.

After this step, we heuristically added special symbols frequently used in user-generated Korean textual data as well as several other languages (Latin alphabet, Kana, Kanji etc.) to the vocabulary.

In \autoref{tab:params} we present the vocabulary sizes for our model, comparing with multilingual BERT and several previous studies on Korean pre-trained BERT. KR-BERT has a vocabulary of 16424 character tokens consisting of both single characters and concatenated characters as a subword unit. KR-BERT’s sub-character vocabulary has 12356 tokens which were constructed in the same manner. Our vocabulary size is much smaller than multilingual BERT and KorBERT and similar to KoBERT.

\subsection{Subword Tokenization}
We used the WordPiece tokenization \cite{wu2016google} and our BidirectionalWordPiece tokenization.

\subsubsection{Baselines}
The baselines we used are the original WordPiece tokenization and the SentencePiece tokenization \cite{kudo-2018-subword}. As in the original BERT model, the WordPiece algorithm was used for tokenization. This resulted in the original corpus being segmented into subword level tokens. Multilingual BERT and KorBERT also used this algorithm. On the other hand, KoBERT used the SentencePiece algorithm as a comparison of subword tokenization methods.

The WordPiece Tokenizer used by multilingual BERT adopts the Byte-Pair-Encoding (BPE) \cite{sennrich-etal-2016-neural} algorithm. This algorithm is deterministic, breaking down a sentence into one unique sequence. In contrast, the SentencePiece Tokenizer uses a Unigram Language Model. This probabilistic model investigates all possible segmentations and selects the one with the highest probability.

\subsubsection{BidirectionalWordPiece Tokenizer}

We use the BidirectionalWordPiece model to reduce search costs while maintaining the possibility of choice. This model applies BPE in both forward and backward directions to obtain two candidates and chooses the one that has a higher frequency.

In the Korean language, typical noun phrases and verb phrases form in different ways. In a noun declension, a long stem is followed by one or two short particles. On the contrary, a verb conjugation has a short stem, to which multiple endings are attached. In the case of noun phrases, it is advantageous to match the longer subword unit from the left, while the backward match is more appropriate for verb phrases, hence our use of a bidirectional tokenizer.

Our tokenizer is designed to tokenize the data with the corresponding vocabulary, depending on the data representation: character or sub-character.

\subsection{Comparison with Other Korean Models}

We compare our KR-BERT with Multilingual BERT, KorBERT, and KoBERT.

\autoref{tab:params} shows the comparison of the vocabulary size, the number of parameters, and the size of the training data between the models. KorBERT has a large vocabulary, many parameters, and a large data size which results in a high memory requirement and longer training time. Meanwhile, KoBERT has a smaller vocabulary and fewer parameters than KR-BERT, but our model has a smaller data size.

\begin{table}[t]
    \makegapedcells
    \centering
    \resizebox{\columnwidth}{!}{
    \begin{tabular}{l|ccccc}  
    \toprule
    & \textbf{\makecell{Multilingual \\ BERT}} & \textbf{\makecell{KorBERT}} & \textbf{KoBERT} & \textbf{\makecell{KR-BERT \\ character}} & \textbf{\makecell{KR-BERT \\ sub-character}} \\
    \midrule
    \textbf{vocabulary size} & 119547 & 30797 & 8002 & 16424 & 12367 \\
    \textbf{parameter size} & 167,356,416 & 109,973,391 & 92,186,880 & 99,265,066 & 96,145,233 \\
    \textbf{data size} & \makecell{- \\ (The Wikipedia data \\ for 104 languages)} & \makecell{23GB \\ 4.7B morphemes} & \makecell{- \\ 25M sentences, \\ 324M words} & \makecell{2.47GB \\ 20M sentences, \\ 233M words} & \makecell{2.47GB \\ 20M sentences, \\ 233M words} \\
    \bottomrule
    \end{tabular} 
    }
    \caption{Comparison of the vocabulary size, the number of parameters, and the training data size between the models trained on Korean texts}
    \label{tab:params}
\end{table}

\autoref{tab:vocabs}  represents the vocabulary composition of each model. As the proportion of words and subwords in Hangul shows, a Korean-specific model such as KorBERT, KoBERT and our KR-BERT is perferable to the multilingual BERT. 

\begin{table}[t]
    \makegapedcells
    \centering
    \resizebox{\columnwidth}{!}{
    \begin{tabular}{l|ccccc}  
    \toprule
    & \textbf{\makecell{Multilingual \\ BERT}} & \textbf{\makecell{KorBERT}} & \textbf{KoBERT} & \textbf{\makecell{KR-BERT \\ character}} & \textbf{\makecell{KR-BERT \\ sub-character}} \\
    \midrule
    \textbf{\makecell[l]{words \\ (Hangul)}} & \makecell{1664 \\ (1.391\%)} & \makecell{12047 \\ (39.117\%)} & \makecell{4489 \\ (56.098\%)} & \makecell{7352 \\ (44.764\%)} & \makecell{6606 \\ (53.416\%)} \\
    \textbf{\makecell[l]{subwords \\ (Hangul)}} & \makecell{1609 \\ (1.346\%)} & \makecell{8023 \\ (26.051\%)} & \makecell{2678 \\ (33.467\%)} & \makecell{3840 \\ (23.380\%)} & \makecell{2140 \\ (17.304\%)} \\
    \textbf{\makecell[l]{symbols and \\ other languages}} & \makecell{116170 \\ (97.175\%)} & \makecell{10720 \\ (34.808\%)} & \makecell{830 \\ (10.372\%)} & \makecell{5227 \\ (31.825\%)} & \makecell{3616 \\ (29.239\%)} \\
    \textbf{special tokens} & \makecell{5 \\ (0.004\%)} & \makecell{7 \\ (0.023\%)} & \makecell{5 \\ (0.062\%)} & \makecell{5 \\ (0.030\%)} & \makecell{5 \\ (0.040\%)} \\
    \midrule
    \textbf{total} & 119547 & 30797 & 8002 & 16424 & 12367 \\
    \bottomrule
    \end{tabular} 
    }
    \caption{Vocabulary composition of the models trained on Korean texts}
    \label{tab:vocabs}
\end{table}

\begin{table*}[tb]
    \centering
    \makegapedcells
    \resizebox{\textwidth}{!}{
    \begin{tabular}{l|ccccccc}  
    \toprule
    & \textbf{\makecell{Multilingual \\ BERT}} & \textbf{\makecell{KorBERT \\ character}} & \textbf{KoBERT} & \textbf{\makecell{KR-BERT \\ character \\ WordPiece}} & \textbf{\makecell{KR-BERT \\ character \\ BidirectionalWordPiece}} & \textbf{\makecell{KR-BERT \\ sub-character \\ WordPiece}} & \textbf{\makecell{KR-BERT \\ sub-character \\ BidirectionalWordPiece}}\\
    \midrule
    \textbf{\makecell{nayngcangko \\ ``refrigerator''}} & \makecell{nayng\#cang\#ko} & \makecell{nayng\#cang\#ko} & \makecell{nayng\#cang\#ko} & \makecell{nayngcangko} & \makecell{nayngcangko} & \makecell{nayngcangko} & \makecell{nayngcangko} \\
    \textbf{\makecell{chwupta \\ ``cold''}} & \makecell{[UNK]} & \makecell{chwup\#ta} & \makecell{chwup\#ta} & \makecell{chwup\#ta} & \makecell{chwup\#ta} & \makecell{chwu\#pta} & \makecell{chwu\#pta} \\
    \textbf{\makecell{paytsalam \\ ``seaman''}} & \makecell{[UNK]} & \makecell{payt\#salam} & \makecell{payt\#salam} & \makecell{payt\#salam} & \makecell{payt\#salam} & \makecell{pay\#t\#salam} & \makecell{pay\#t\#salam} \\
    \textbf{\makecell{maikhu \\ ``microphone''}} & \makecell{ma\#i\#khu} & \makecell{mai\#khu} & \makecell{ma\#i\#khu} & \makecell{maikhu} & \makecell{maikhu} & \makecell{maikhu} & \makecell{maikhu} \\ 
    \bottomrule
    \end{tabular} 
    }
    \caption{Tokenization Examples of Korean-specific models}
    \label{tab:tokens}
\end{table*}

\autoref{tab:tokens} compares the tokenization (segmentation) results. Multilingual BERT often results in the [UNK] token. Tokens from KR-BERT models retain the original meanings well, proving the superiority of our tokenizer and vocabularies. Frequent word 냉장고 nayngcangko `refrigerator' is only tokenized by KR-BERT. It seems that the heuristically added vocabularies and sub-character level segmentation of our models improve the tokenization. Furthermore, KR-BERT models are more robust when it comes to foreign loan words (마이크 maikhu `microphone') and, especially for sub-character models, conjugation. For example, 춥다 chwupta `cold' is segmented as its conjugation form 추 chwu and ㅂ다 pta in our KR-BERT sub-character model. 뱃사람 paytsalam `seaman' is segmented as 배 pay `boat' and 사람 salam `man' with ㅅ t between the tokens due to a phonological rule.

\section{Experiments and Results}

Our experiments include pre-training the BERT model on data obtained from Korean Wikipedia and news articles, followed by running experiments on the downstream tasks of sentiment classification, question answering, named entity recognition and paraphrase detection. The results are compared against the baseline of the pre-trained multilingual BERT model and the existing Korean-only BERT models.

\subsection{Dataset and Preprocessing}

For data preprocessing, we extracted text from the Korean Wikipedia dump using WikiExtractor, omitting metadata and history. We also extracted crawled news articles, such as Chosun Ilbo. After this, we tokenized all the text into sentences using the Natural Language Toolkit (NLTK) \cite{loper2002nltk}.

For our sub-character version, we decomposed all Hangul syllable characters into their sub-characters using Unicode’s Normalization Form D.

\subsection{Results}
\subsubsection{Masked LM Accuracy}
We present the training results of the masked language models in \autoref{tab:mlm}. The masked LM accuracy of multilingual BERT and KorBERT were not specified. We observe that our character based model and sub-character based model have higher accuracy than KoBERT, and when using a BidirectionalWordPiece tokenizer we can also get slightly better language model accuracy.

\begin{table}[t]
    \centering
    \resizebox{\columnwidth}{!}{
    \begin{tabular}{lr}  
    \toprule
    \textbf{Model} & \textbf{Masked LM Accuracy} \\
    \midrule
    \textbf{Multilingual BERT}& n/a \\[0.05cm]
    \textbf{KorBERT} & n/a \\[0.05cm]
    \textbf{KoBERT} & 0.750 \\[0.05cm]
    \textbf{\makecell[l]{KR-BERT character WordPiece}} & 0.773 \\[0.05cm]
    \textbf{\makecell[l]{KR-BERT character BidirectionalWordPiece}} & \textbf{0.779} \\[0.05cm]
    \textbf{\makecell[l]{KR-BERT sub-character WordPiece}} & 0.761 \\[0.05cm]
    \textbf{\makecell[l]{KR-BERT sub-character BidirectionalWordPiece}} & 0.769 \\
    \bottomrule
    \end{tabular} 
    }
    \caption{Masked LM Accuracy}
    \label{tab:mlm}
\end{table}

\subsubsection{Downstream tasks}
We tested sentiment classification, question answering,named entity recognition and paraphrase detection using Naver Sentiment Movie Corpus, KorQuAD, KorNER, and Korean Paired Question dataset. 

We use a batch size of 32 and 64 and fine-tuned for 5 epochs over the data for all tasks. For each task, we selected the best batch size, the best epochs, and the best fine-tuning learning rate (among 1e-3, 2e-5, 5e-5) on the dev set. 

\begin{table*}[t]
    \centering
    \resizebox{\textwidth}{!}{
    \begin{tabular}{lcccc}  
    \toprule
    \textbf{Model} & \textbf{NSMC (Acc.)} & \textbf{KorQuAD (F1)} & \textbf{KorNER (F1)} & \textbf{\makecell{Paraphrase \\ Detection (Acc.)}} \\ 
    \midrule
    \textbf{\makecell[l]{Multilingual BERT (Google)}} & 87.08 & 89.58 & 61.52 & 79.55 \\
    \textbf{\makecell[l]{KorBERT (ETRI)}} & \textbf{89.84} & 83.73 & 59.43 & \textbf{93.79} \\
    \textbf{\makecell[l]{KoBERT (SKT)}} & 89.01 & n/a & n/a & 91.03 \\
    \midrule 
    \textbf{\makecell[l]{KR-BERT character WordPiece}} & 89.34 & \textbf{89.92} & 64.97 & \textbf{93.54} \\
    \textbf{\makecell[l]{KR-BERT character BidirectionalWordPiece}} & \textbf{89.38} & 89.18 & 64.50 & 92.74 \\

    \textbf{\makecell[l]{KR-BERT sub-character WordPiece}} & 89.20 & 89.31 & \textbf{66.64} & 93.14 \\
    \textbf{\makecell[l]{KR-BERT sub-character BidirectionalWordPiece}} & 89.34 & 89.78 & 66.28 & 92.74 \\
    \bottomrule
    \end{tabular} 
    }
    \caption{Performances on Downstream Tasks}
    \label{tab:results}
\end{table*}

Results are presented in \autoref{tab:results}. KR-BERT performs comparably or better than the other Korean based model, and it has more stable higher results than others.

In the NSMC task, while KorBERT shows the highest accuracy, all the KR-BERT models more consistently show high accuracies compared to KoBERT and multilingual BERT. KR-BERT character WordPiece model, followed by sub-character BidirectionalWordPiece model, achieves the highest F1 score at the KorQuAD. At the KorNER task, all the KR-BERT sub-character models appear to be the best models with the original WordPiece tokenizer model recording the highest and our BidirectionalWordPiece tokenizer model following the second highest. KR-BERT character models had the next highest score, displaying big differences between KR-BERT and the other models. On the Paraphrase detection task, KorBERT is the most accurate model, but only by a narrow margin and the KR-BERT character WordPiece model reveals the second highest accuracy.

Although for NSMC and Paraphrase Detection, our KR-BERT model has lower results than the KorBERT model, the results are comparable and it still obtains about 7\% improvement over the KorBERT model for KorQuAD and KorNER.

\subsection{Analysis of Downstream Tasks}


\begin{table*}[t]
    \centering
    \makegapedcells
    \resizebox{\textwidth}{!}{
    \begin{tabular}{l|ccccccc}  
    \toprule
    & \textbf{\makecell{Multilingual \\ BERT}} & \textbf{\makecell{KorBERT \\ character}} & \textbf{KoBERT} & \textbf{\makecell{KR-BERT \\ character \\ WordPiece}} & \textbf{\makecell{KR-BERT \\ character \\ BidirectionalWordPiece}} & \textbf{\makecell{KR-BERT \\ sub-character \\ WordPiece}} & \textbf{\makecell{KR-BERT \\ sub-character \\ BidirectionalWordPiece}}\\[0.05cm]
    \midrule 
    \textbf{\makecell{iyenghwa \\ ``thismovie''}} & \makecell{i\#yeng\#hwa \\ (incorrect)} & \makecell{iyenghwa \\ (incorrect)} & \makecell{i\#yengwha \\ (correct)} & \makecell{iyeng\#hwa \\ (incorrect)} & \makecell{i\#yenghwa \\ (correct)} & \makecell{iyeng\#hwa \\ (incorrect)} & \makecell{i\#yenghwa \\ (correct)} \\[0.25cm] 
    \textbf{\makecell{caymisnunteyng \\ ``It's amusing''}} & \makecell{[UNK] \\ (incorrect)} & \makecell{cay\#mis\#nun\#teyng \\ (incorrect)} & \makecell{cay\#mis\#nun\#teyng \\ (incorrect)} & \makecell{cay\#mis\#nun\#teyng \\ (incorrect)} & \makecell{cay\#mis\#nun\#teyng \\ (incorrect)} & \makecell{caymi\#s\#nuntey\#ng \\ (correct)} & \makecell{caymi\#s\#nuntey\#ng \\ (correct)} \\[0.1cm]
    \bottomrule
    \end{tabular} 
    }
    \caption{Tokenization Examples of Korean-specific models}
    \label{tab:examples}
\end{table*}

For all of the KR-BERT models, the character WordPiece model achieved remarkable results on all of the downstream tasks. Our new KR-BERT sub-character BidirectionalWordPiece model does not perform as well but still shows good results consistently, especially for the NSMC and NER tasks. The BidirectionalWordPiece tokenizer models display higher accuracy for the NSMC, and the NER the F1 score from its sub-character models are the highest. We emphasize that NSMC, the dataset for sentiment classification, is noisy user-generated data consisting of real comments from users. Additionally, the NER task contains a lot of out-of-vocabulary (OOV) words, mostly consisting of low-frequency noun stems. The high performance of KR-BERT with the BidirectionalWordPiece tokenizer shows that this tokenizer and sub-character level representation are able to effectively tackle the OOV problem, or the incorrect spacing problem. An example of this from the NSMC dataset can be seen in \autoref{tab:examples}.

\raggedbottom

Like other movie review data, the NSMC contains various types of non-standard language use, including deliberate spelling errors exemplified in \autoref{tab:examples}. The phrase, 재밋는뎅 cay.mis.nun.teyng, whose standard form is 재밌는데 cay.miss.nun.tey ``It’s amusing,'' should be easily classified if tokenized properly to capture the meaningful words 재미 caymi ``amusing'' and 는데 nuntey ``it’s,'' as in the KR-BERT sub-character models are able to do. 

Because NER and paraphrase detection consists of relatively formal data, models using the WordPiece tokenizer have higher scores than models using BidirectionalWordPiece tokenizer.

\begin{table}[t]
    \centering
    \makegapedcells
    \resizebox{\columnwidth}{!}{
    \begin{tabular}{lrrr}  
    \toprule
    \textbf{Model} & \textbf{\# of [UNK]} & \textbf{total \# of tokens} & \textbf{[UNK] ratio} \\
    \midrule
    \textbf{Multilingual BERT} & 3461 & 337741 & 0.01024 \\[0.05cm]
    \textbf{KorBERT} & 16 & 300437 & 0.00005 \\[0.05cm]
    \textbf{\makecell[l]{KR-BERT character \\ WordPiece}} & 2336 & 293418 & 0.00796 \\
    \textbf{\makecell[l]{KR-BERT character \\ BidirectionalWordPiece}} & 2336 & 293441 & 0.00796 \\
    \textbf{\makecell[l]{KR-BERT sub-character \\ WordPiece}} & 45 & 293266 & 0.00015 \\
    \textbf{\makecell[l]{KR-BERT sub-character \\ BidirectionalWordPiece}} & 45 & 293467 & 0.00015 \\
    \bottomrule
    \end{tabular} 
    }
    \caption{[UNK] ratio of each model for KorNER data}
    \label{tab:unk}
\end{table}

Especially in NER which has more low frequency proper nouns than other datasets, sub-character based models get much higher F1 scores because they can capture the unknown words more effectively. To look into the OOV problem in more detail, \autoref{tab:unk} shows the [UNK] ratio for each model, representing the number of words tokenized as [UNK] divided by the total number of words. 

Because KorQuAD and paraphrase detection consists of relatively formal data, models using the WordPiece tokenizer have much higher scores than models using the BidirectionalWordPiece tokenizer.

\section{Conclusion}

We have demonstrated that our Korean-specific KR-BERT model can effectively deal with Korean NLP tasks. The KR-BERT model shows a higher performance than multilingual BERT in both Masked LM accuracy and downstream tasks, which means it captures language-specific linguistic phenomena. Compared to other Korean models, KR-BERT has higher or comparable results in downstream tasks. We introduced a sub-character based model and the BidirectionalWordPiece tokenizer to deal with a morphologically rich language with poor resources, and can confirm that a sub-character based Korean model and BidirectionalWordPiece tokenizer have efficacy in certain datasets. Our contribution was to test a sub-character based Korean model using its morphological features and the BidirectionalWordPiece tokenizer with a small pre-training corpus. For future work, we plan to continue making much smaller and faster Korean language models by combining current approaches with distilling methods.

\bibliographystyle{named}
\bibliography{krbert}

\end{document}